%% file: EACL2021-RJ.tex
\documentclass[11pt,a4paper]{article}
\usepackage[hyperref]{eacl2021}
\usepackage{times}
\usepackage{latexsym}
\usepackage{graphicx}
\usepackage{xspace}
\usepackage{multirow, makecell}
\usepackage{amsmath}
\usepackage{amsthm,bm}
\usepackage{microtype}
\usepackage{subfiles}
\usepackage{subcaption}
\usepackage{lipsum}
\usepackage{multirow}
\usepackage{arydshln}
\usepackage{booktabs}
\usepackage{amsfonts}
\usepackage{cleveref}
\usepackage{todonotes}
\usepackage{enumitem}
\usepackage{amssymb}
\usepackage{graphicx}
\usepackage{microtype}
\usepackage{wasysym}
\usepackage{makecell}
\usepackage{todonotes}

\makeatletter
\newcommand*\bigcdot{\mathpalette\bigcdot@{.5}}
\newcommand*\bigcdot@[2]{\mathbin{\vcenter{\hbox{\scalebox{#2}{$\m@th#1\bullet$}}}}}
\makeatother

\crefformat{section}{Section~#2#1#3} %
\crefformat{subsection}{Section~#2#1#3}
\crefformat{subsubsection}{Section~#2#1#3}
\crefformat{section}{Section~#2#1#3}
\crefname{table}{{T}able~}{{T}ables~}
\Crefname{table}{Table~}{Tables~}
\crefname{figure}{{F}igure~}{{F}igures~}
\Crefname{figure}{Figure~}{Figures~}

\newcommand{\concat}{\ensuremath{\oplus}}

\newcommand{\astwo}{$\text{AS2}$\xspace}
\newcommand{\robertabase}{$\text{RoBERTa}_\textsc{base}$\xspace}
\newcommand{\gpd}{$\text{WQA}$\xspace}

\definecolor{ylw}{HTML}{B29334}
\definecolor{ble}{HTML}{587BB0}
\definecolor{rdd}{HTML}{923936}
\definecolor{prp}{HTML}{755386}

\definecolor{cand}{HTML}{C07707}
\definecolor{local}{HTML}{308B56}
\definecolor{glob}{HTML}{C00001}

\aclfinalcopy %
\def\aclpaperid{1142} %

\title{Modeling Context in Answer Sentence \\ Selection Systems on a Latency Budget}

\author{
  Rujun Han\thanks{\hspace{1em}Work was conducted while the author was an intern at Amazon Alexa.} \\
  University of Southern California  \\ Los Angeles, CA, USA \\
  \texttt{rujunhan@usc.edu} \\\And
  Luca Soldaini \\
  Amazon Alexa  \\ Manhattan Beach, CA, USA \\
  \texttt{lssoldai@amazon.com} \\\And
  Alessandro Moschitti \\
  Amazon Alexa  \\ Manhattan Beach, CA, USA \\
  \texttt{amosch@amazon.com} \\}

\date{}

\begin{document}
\maketitle

\subfile{sections/abstract}

\subfile{sections/intro}

\subfile{sections/methodology}

\subfile{sections/results}

\subfile{sections/conclusions}

\bibliography{EACL2021,aaai,cascade}
\bibliographystyle{acl_natbib}

\end{document}

%% file: sections/abstract.tex
\begin{abstract}
    Answer Sentence Selection (\astwo) is an efficient approach for the design of open-domain Question Answering (QA) systems.
    In order to achieve low latency, traditional \astwo models score question-answer pairs individually, ignoring any information from the document each potential answer was extracted from.
    In contrast, more computationally expensive models designed for machine reading comprehension tasks typically receive one or more passages as input, which often results in better accuracy.
    In this work, we present an approach to efficiently incorporate contextual information in \astwo models.
    For each answer candidate, we first use unsupervised similarity techniques to extract relevant sentences from its source document, which we then feed into an efficient transformer architecture fine-tuned for \astwo.
    Our best approach, which leverages a multi-way attention architecture to efficiently encode context, improves 6\% to 11\% over non-contextual state of the art in \astwo with minimal impact on system latency.
    All experiments in this work were conducted in English. 
\end{abstract}

%% file: sections/intro.tex
\section{Introduction}
\label{sec:intro}

\astwo models for open-domain QA typically consider sentences from webpages as independent candidate answers for a given question.
For any webpage containing potential answer candidates for a question, \astwo models first extract individual sentences, then independently estimate their likelihood of being correct answers;
this approach enable highly efficient processing of entire documents.
However, under this framework, context information from the entire webpage (global context), which could be crucial for selecting correct answers, is ignored.
Conversely, current systems in Machine Reading (MR) \cite{huang2019cosmos,NQ-2019,lee-etal-2019-latent,joshi-etal-2020-spanbert} uses a much larger context from the retrieved documents.
MR models receive a question and one or more passages retrieved through a search engine as input; 
they then select one or more spans from the input passages to return as answer.

While being potentially more accurate, MR models typically have higher computational requirements (and thus higher latency) than AS2 models.
That is because MR models need to process passages in their entirety before an answer can be extracted; 
conversely, AS2 systems break down paragraphs in candidate sentences, and evaluate them all at once in parallel.
Therefore, in many practical applications, MR models are only used to examine 10 to 50 candidate passages; 
in contrast, \astwo approaches can potentially process hundreds of documents, e.g., \cite{yoshitomo2020reranking,soldaini2020cascade}.

In this work, we study techniques that can combine the efficacy of MR models with the efficiency of \astwo approaches, while keeping a single sentence as target answer, as in related \astwo works\footnote{MR systems have a different aim than \astwo.}.
In particular, we focus our efforts on improving accuracy of \astwo systems without affecting their latency.

\begin{figure}[t]
    \includegraphics[width=\columnwidth]{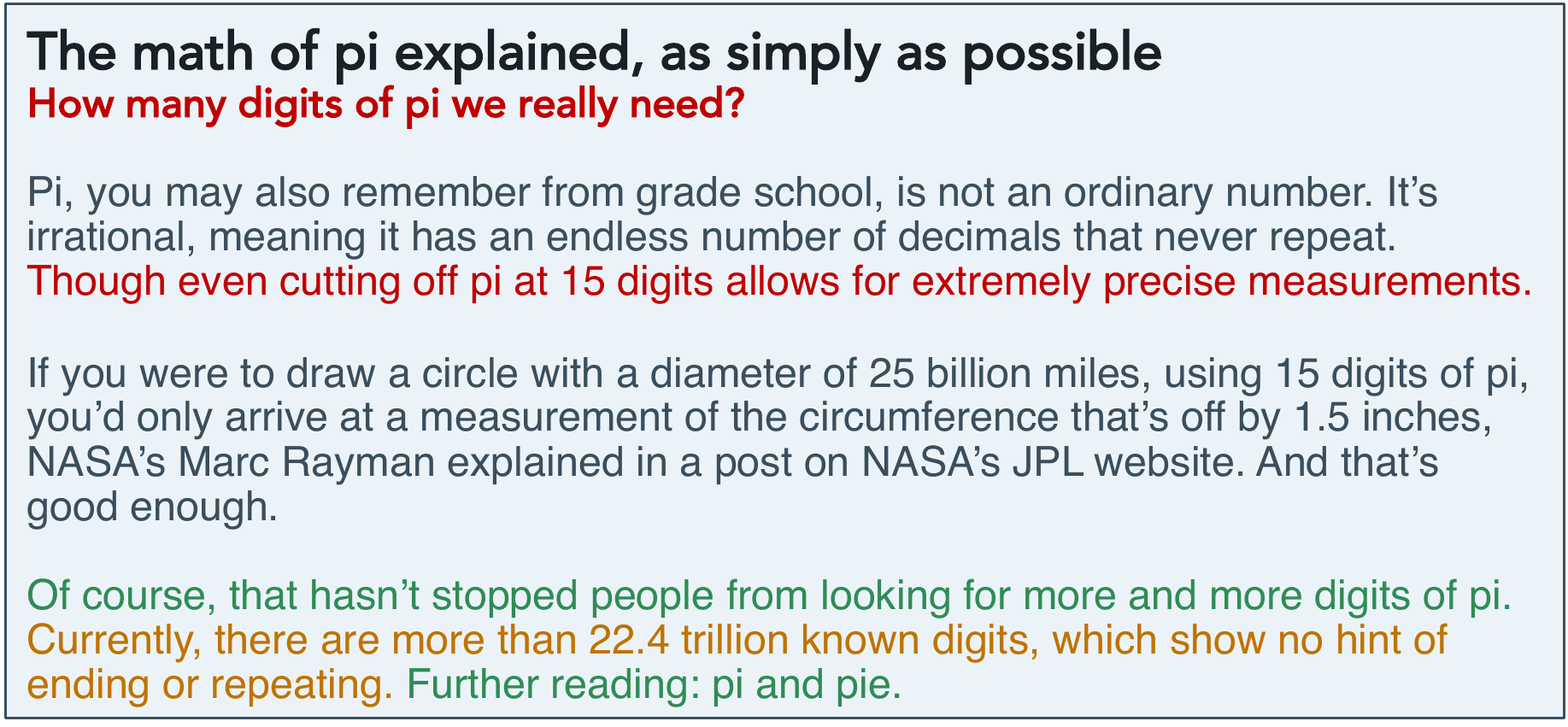}
    \caption{Candidate sentence (in {\color{cand} orange}) for question ``\textit{how many digits are in pi?}''.
    Local context is shown in {\color{local} green}, while global context is shown in {\color{glob} red}.
    }
    \label{fig:context}
\end{figure}

Early neural models for retrieval-based QA focused on incorporated neighboring sentences (local context) to improve performance.
For example, \citet{tan2017context} proposed a neural architecture based on gated recurrent units to encode question, answer, and local context; their approach, while effective at the time, shows a significant gap to the current state-of-the-art models \citep{garg2020tanda}.
\citet{min2018efficient} studied neural efficient models for MR by optimizing answer candidate extraction.
More recently, researchers have focused in including source document information in transformer models.
For example, \citet{joshi2020contextualized} proposed a contextualized model for MR that augments named entities in candidate passages with snippets extracted from Wikipedia pages. Their approach, while interesting, is limited to entities-based context, and specific to Wikipedia and MR domain.
For \astwo, \citet{lauriola2020context} proposed a model that uses local context as defined  by the preceding and following sentences of the target answer.
They also introduced a simple bag-of-words representation of documents as global context, which did not show significant improvement over non-contextual \astwo models.

Unlike previous efforts, our approaches consider both \textit{local context} (that is, the sentences immediately preceding or succeeding a candidate answer), as well as \textit{global context} (phrases in documents that represent the overall topics of a page), as they can both uniquely contribute to the process of selecting the right answer.
As shown in the example in \cref{fig:context}, local context can help disambiguate cases where crucial entities are not present in the candidate answer (there's no mention of \textit{``pi''} in \textit{``[c]urrently, there are more than 22.4 trillion known digits''});
conversely, global context can help reaffirm the relevance of a candidate answer in cases where noisy information is extracted as local context (in the example, \textit{``[f]urther reading: pi and pie''} does not contain any relevant information).

The contributions of this work are:
(\textbf{i}) first, we introduce two effective techniques to extract relevant local and global contexts for a given question and candidate answer;
(\textbf{ii}) then, we propose three different methods for combining contextual information for \astwo tasks;
(\textbf{iii}) finally, we evaluate our approaches on two \astwo datasets: ASNQ \cite{garg2020tanda} and a benchmark dataset we built to evaluate real-world QA systems.
Results show that our most efficient system, which leverages a multi-way attention architecture, can improve over the previous non-contextual state of the art model for \astwo by up to 11\%;
furthermore, these results are achieved while maintaining similar efficiency to the best-performing, non-contextual \astwo systems, making our approach a viable strategy for latency-sensitive applications.
Code and models are made available at \texttt{\url{https://github.com/alexa/wqa-contextual-qa}}.

%% file: sections/methodology.tex
\section{Methodology}
\label{sec:method}

\begin{table}
\centering
\renewcommand{\arraystretch}{1.15}
\small
\begin{tabular}{p{22em}}
\toprule
\textbf{Question}: ``where did the potter's wheel first develop'' \\
\textbf{Corrent Answer}: ``Tournettes, in use around 4500 BC in the Near East, were turned slowly by hand or by foot while coiling a pot'' \\
\textbf{Sentence selected by N-grams}: ``In the Iron Age, the potter 's wheel in common use had a turning platform about one metre (3 feet) above the floor, connected by a long axle to a heavy flywheel at ground level. Use of the potter's wheel became widespread throughout the Old World but was unknown in the Pre-Columbian New World, where pottery was handmade by methods that included coiling and beating.'' \\
\midrule
\textbf{Question}: ``where do pineapples come from in the world'' \\
\textbf{Correct answer}: ``In 2016, Costa Rica, Brazil, and the Philippines accounted for nearly one-third of the world's production of pineapple.'' \\
\textbf{Sentence selected by Cosine Similarity}: ``The plant is indigenous to South America and is said to originate from the area between southern Brazil and Paraguay''\\
\bottomrule
\end{tabular}
\caption{Examples of \textbf{global context} selected via N-gram similarity (top) and cosine similarity (bottom).
Overall, the N-gram approach tends to select longer context sentences than Cosine’s, which in turn leads to fewer context sentences being included in the global context (as we limit it to 128 tokens). 
Empirically, we also noticed that N-gram selected context sentences also contain more noise.
}
\label{table:global_example}
\end{table}

Our approach to ranking candidate answer consists of two components: the first (\cref{sec:method:context}) is responsible for extracting context for each candidate answer, while the second (\cref{sec:method:models}) encodes information from local and global contexts to score each question / candidate answer pair.

\subsection{Context Construction}
\label{sec:method:context}

\begin{figure*}
  \begin{subfigure}[b]{.32\textwidth}
  \centering
  \includegraphics[height=11em]{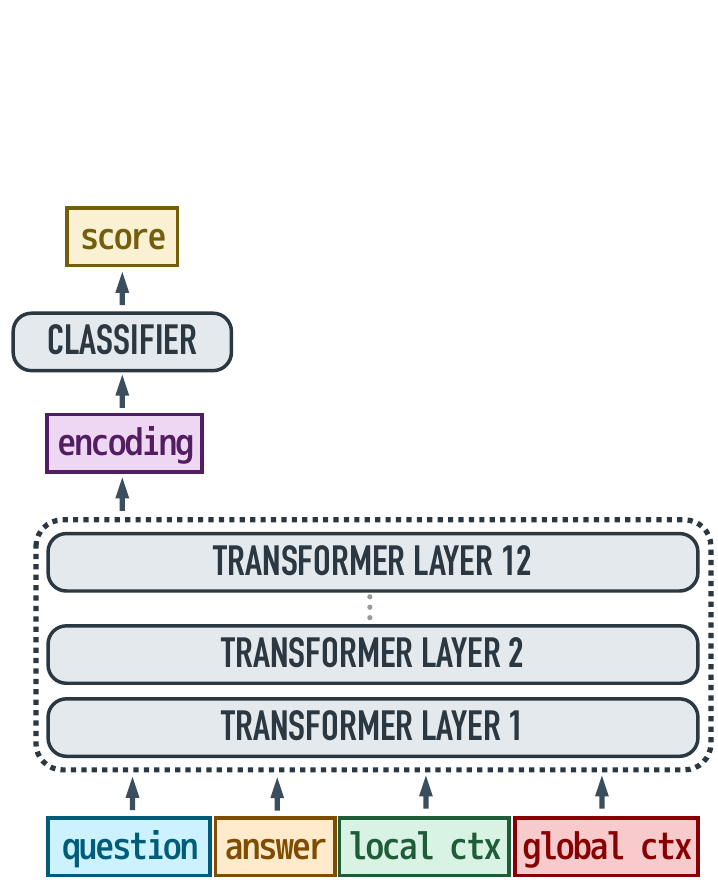}
  \caption{}\label{fig:concat}
  \end{subfigure}
  \hfill
  \begin{subfigure}[b]{.32\textwidth}
  \centering
  \includegraphics[height=11em]{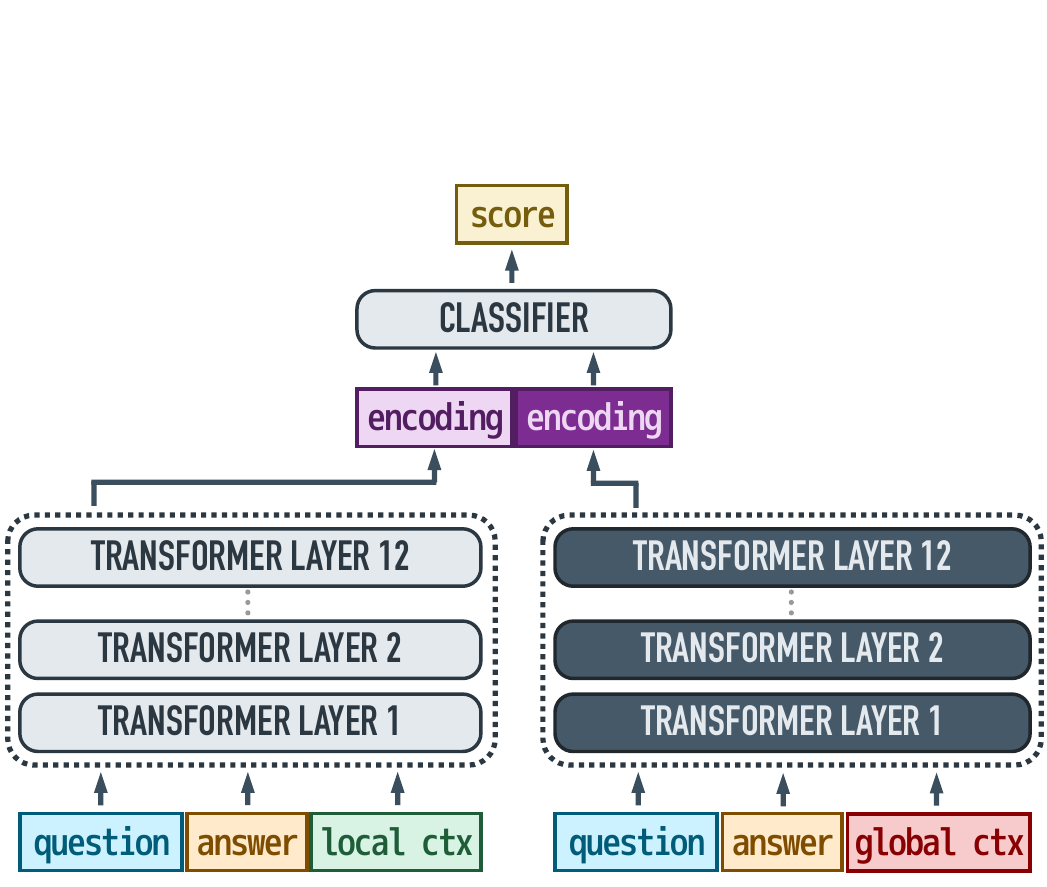}
  \caption{}\label{fig:ensemble}
  \end{subfigure}
  \hfill
  \begin{subfigure}[b]{.32\textwidth}
  \centering
  \includegraphics[height=11em]{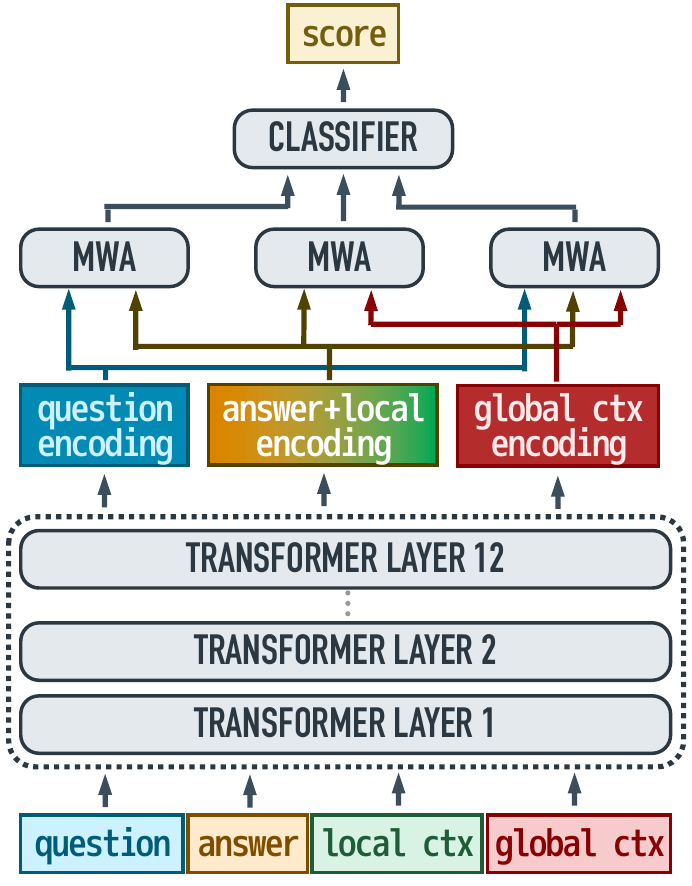}
  \caption{}\label{fig:mwa}
  \end{subfigure}
  \vspace{-0.5em}
  \caption{From left to right, the three approaches we evaluated in this work: context concatenation (\cref{fig:concat}), context ensemble (\cref{fig:ensemble}), and multi-way attention (\cref{fig:mwa}).}
  \vspace{-0.5em}
  \label{fig:models}
\end{figure*}

As previously mentioned, our proposed method for contextualizing answers relies on enriching them with information encoded in sentences adjacent to them, as well as from sentences throughout the document each potential answer comes from; we will define these extraction processes in this section.

In the rest of this work, we will use $Q$ to refer to a question and $\mathcal{D} = \left\{D_1, \ldots, D_i, \ldots, D_N\right\}$ to indicate a collection of documents containing potential answers for $Q$.
Each document $D_i$ is comprised of an ordered sequence of sentences $D_i = \langle C_{i,1}, \ldots, C_{i,j}, \ldots, C_{i, M} \rangle$; each sentence $C_{i,j}$ could be used either as candidate answer, or as context for another candidate.

\subsubsection{Local Context}
\label{sec:method:context:local}

Similarly to previous work \citep{tan2017context,lauriola2020context}, we define local context $\text{Loc}_k(C_{i,j})$ for candidate $C_{i, j}$ as the sentences immediately preceding and succeeding each answer candidate within a window of $2k + 1$ sentences, i.e., $\text{Loc}_k(C_{i,j}) = \langle C_{i,j - k}, \ldots C_{i,j-1}, C_{i,j + 1} \ldots C_{i, j + k} \rangle$.
In our experiments, we tried constructing a local context of up to six sentences; however, we observed diminishing return when using more than the previous and next sentences (i.e., $k=1$) at the expense of more computational complexity.
Therefore, the results presented in this work use two adjacent sentences as local context.

\subsubsection{Global Context}
\label{sec:method:context:global}

Unlike local context, there are many potential approaches to extracting information that can be used to understand relevancy of a candidate answer to a question.
We proposed and evaluated two different techniques for extracting global context $\text{Glo}_h(C_{i,j})$ (examples for both are shown in Table~\ref{table:global_example}).

\paragraph{N-gram Overlap}
Similarly, to \citet{joshi2020contextualized}, we experimented with selecting sentences as global context based on their n-gram overlap with question and candidates.

In detail, we first extract the set of  all unigrams, bigrams, and trigrams from question $Q$ and candidate $C_{i,j}$, which we denote as $\text{Ng}_{1,2,3}(Q, C_{i,j})$;
then, we repeat the same procedure for all $\{C_{i, p} \subset \mathcal{D} \text{ where } p \neq j\}$ to obtain $\text{Ng}_{1,2,3}(C_{i,p})$.
Finally, we score each sentence as follows:\vspace{-1.6em}\\
\begin{multline}
    \text{Score}_{ngrams}(C_{i,p} | Q, C_{i,j}) = \\
    \frac{|\text{Ng}_{1,2,3}(C_{i,p}) \cap  \text{Ng}_{1,2,3}(Q, C_{i,j})|}{|\text{Ng}_{1,2,3}(Q, C_{i,j})|}
\end{multline}\vspace{-.3em}
and pick the top $h$ sentences as global context.

\paragraph{Semantic Similarity}
N-grams overlap can only extract spans of text that are lexically similar to either the query or candidate.
To better capture context that is topically relevant to an answer, we also propose to use cosine distance between sentence embeddings to approximate semantic similarity.

Given a sentence encoder model\footnote{In our experiments, we use non-finetuned \robertabase model \citep{liu2019roberta}.} $\mathcal{M}$, we first obtain a representations for the question-answer pair $\mathcal{M}(Q\concat C_{i,j})$ and context sentences $\{\mathcal{M}(C_{i,p}) \text{ for all } p=\{1,\ldots, M\}, p \neq j$; then we pick the top $h$ sentences maximizing the following cosine similarity score as global context:
\begin{multline}
    \text{Score}_{sim}(C_{i,p} | Q, C_{i,j}) = \\
    \frac{\mathcal{M}(Q\concat C_{i,j}) \bigcdot{} \mathcal{M}(C_{i,p})}{||\mathcal{M}(Q\concat C_{i,j})|| \hspace{6pt} ||\mathcal{M}(C_{i,p})||},
\end{multline}
where $\concat$ indicates string concatenation.

\subsection{Contextual \astwo Models}
\label{sec:method:models}

Once local context $\text{Loc}_k(C_{i,j})$ and global context $\text{Glo}_h(C_{i,j})$ are extracted for candidate $C_{i,j}$, we encode them in conjunction with candidate answer and question to estimate the likelihood of $C_{i,j}$ being a correct answer for $Q$.
Our approaches (summarized in \cref{fig:models}) consume up to $h=5$ sentences as global context so not to exceed $128$ tokens. \footnote{Using up 5 sentences resulted in 3.02 (n-gram context) and 2.85 (cosine context) sentences being selected on average for the ASNQ dataset.}
Similarly, to other efforts in this area (e.g., \citep{garg2020tanda}), we leverage state-of-the-art transformer models to estimate said probability.
Specifically, we studied three approaches to encode question and answer context.
Although the methods we proposed can be easily combined with any transformer architecture, all models described here are initialized from a \robertabase checkpoint.

\paragraph{Context Concatenation}
A simple baseline \citep{vanaken2019how,joshi2020contextualized} for encoding multiple contexts is to concatenate each question, candidate answer, and local/global context text and feed them through transformer model (Fig.~\ref{fig:concat});
the resulting encoding is then projected to a probability distribution using a dense feed-forward layer.
This baseline relies on the transformer self-attention mechanism to implicitly model relations between local and global context.

\begin{table}[t]
    \renewcommand{\arraystretch}{1.1}
    \centering
    \small
    \begin{tabular}{@{}lcccc@{}}
    \toprule
    \textbf{Model}          & \textbf{P@1}       & \textbf{MAP}       & \textbf{MRR} & \textbf{Latency}\\
    \midrule
    No Context & \multirow{ 2}{*}{0.596} & \multirow{ 2}{*}{0.685} & \multirow{ 2}{*}{0.706} & \multirow{ 2}{*}{$\mathbf{5.49}$} \\
    {\scriptsize \citep{garg2020tanda}} & & & \\

    Local Context & \multirow{ 2}{*}{0.653} & \multirow{2}{*}{0.732} & \multirow{ 2}{*}{0.752} & \multirow{ 2}{*}{$5.62$} \\
    {\scriptsize \citep{lauriola2020context}} & & & \\
    Global Context & 0.622 & 0.698 & 0.722 & $5.67$ \\
    \midrule
    Context Concat &  0.631 & 0.726 & 0.743 & $5.76$ \\
    Context Ensemble & \textbf{0.668} & \textbf{0.743} & \textbf{0.765} & $7.41$ \\
    MWA & 0.661 & 0.742 & 0.758 & $5.82$ \\
    \bottomrule
\end{tabular}
\caption{Results on the ASNQ Test Set.
The latency is reported in micro-seconds (${\mu}s$) per sample ($\pm 0.1\text{ }{\mu}s$ with 95\% CI).}
\label{tab:asnq}
\end{table}

\paragraph{Context Ensemble}
As mentioned in \cref{sec:intro}, local and global contexts might capture different aspects of the source document of a candidate answer.
To empirically verify this hypothesis, we evaluated an ensemble model that encodes local and global contexts separately using two independent transformer models (\cref{fig:ensemble}).
The two models are independently trained for \astwo;
then, their encodings are concatenated and passed to a feed-forward layer to estimate relevance of candidate $C_{i,j}$ for question $Q$. The top 3 layers\footnote{We tested with approaches for gradual unfreezing of the top $k$ layers; $k=3$ yielded the best validation scores.} model is once again fine tuned on the training set.

\paragraph{Multi-way Attention (MWA)}
While leveraging independent encoders for local and global contexts can lead to an improvement in performance compared to using a single encoder, it also doubles computational requirements.
Therefore, we also explored techniques that incorporate inductive biases into transformer models and achieve efficiency comparable to the context concatenation approach.
One such approach is, as shown in \cref{fig:mwa}, to combine a transformer model with a multi-way attention mechanism \citep{tan2018multiway}, which has been shown to be effective for commonsense reasoning tasks \citep{huang2019cosmos}.
This approach still uses a single transformer model to produce an encoding for a sequence of question, candidate answer, local context, and global context;
however, similarly to the ensemble model, the additional attention mechanism forces the encoder to selectively look at local and global contexts separately.

%% file: sections/results.tex
\section{Experiments}
\label{sec:experiments}

\subsection{Setup}
\label{subsec:setup}

In order to validate the effectiveness of the proposed context modeling techniques, we evaluated our results on two datasets: ASNQ and \gpd.\vspace{-.5em}

\paragraph{ASNQ} The Answer Sentence Natural Questions dataset \citep{garg2020tanda} is a large collection of 59,914 questions and 24,732,396 candidate answers.
It was obtained by extracting sentence candidates from the Google Natural Question (NQ) benchmark~\citep{47761}.
We use the train, development, and test splits proposed by \citet{soldaini2020cascade}.\vspace{-.5em}

\begin{table}[t]
    \centering
    \small
    \renewcommand{\arraystretch}{1.0}
    \begin{tabular}{@{}lccc@{}}
    \toprule
    \textbf{Model}          & \textbf{P@1}       & \textbf{MAP}       & \textbf{MRR}       \\
    \midrule
    No Context   &  \multicolumn{3}{c}{\textit{baseline}} \\
    Global Context & +3.95\% & +2.52\% & +2.29\% \\
    Local Context  & +3.59\% & +2.89\% & +2.21\% \\
    \midrule
    Context Concat & +1.52\% & +2.33\% & +1.34\% \\
    Context Ensemble & \textbf{+5.92\%} & \textbf{+4.10\%} & \textbf{+3.39\%} \\
    MWA & +5.56\% & +3.92\% & +3.08\% \\
    \bottomrule
\end{tabular}
\caption{Results on the \gpd Test Set.}
\label{tab:gpd}
\end{table}

\paragraph{\gpd} The Web-based Question Answering is an in-house dataset built by Alexa AI as part of the effort of understanding and benchmarking QA systems. The creation process includes the following steps:
(i) Given a set of questions, a search engine is used to retrieve up to 100 web pages from an index containing hundreds of million pages.
(ii) From the set of retrieved documents, all candidate sentences are extracted and ranked using \astwo models from \citet{garg2020tanda}; and (iii)
at least 25 candidates for each question are annotated by humans.
Overall, the version of \gpd we used contains 6,962 questions and 283,855 candidate answers. We reserved 3,000 questions for evaluation, 808 for development, and used the rest for training \footnote{The public version of \gpd will be released in the short-term future. Please search for a publication with title \emph{WQA: A Dataset for Web-based Question Answering Tasks} on arXiv.org.}. 

\vspace{1em}
Models were trained on a single machine with 8 NVIDIA Tesla V100 GPUs with 32 GB of memory each. 
We used model implementations from the Transformers library when available \citep{wolf2020transformers}.
All our experiments were computed using mixed precision through NVIDIA apex\footnote{\texttt{\url{https://github.com/NVIDIA/apex}}}. 
Latency was measured on single GPU with a fixed batch size of 128. 
Tokenization and time to transfer tensors to the GPU was not included in the latency values.

\subsection{Results and Discussion}
\label{sec:results}

Results on the ASNQ and \gpd are summarized in tables \ref{tab:asnq} and \ref{tab:gpd}, respectively.
Overall, we observe that the context ensemble model achieves the best performance;
however, as observed in \cref{sec:method:models}, this model is twice as large as a \robertabase model, thus it is a rather expensive solution.

Among our baselines, we note that local context outperforms the model leveraging the global context.
This observation suggests that local information carries more importance in understanding the semantic relationship between question and candidate answers.
Surprisingly, we observe that simply concatenating local and global contexts achieves worse performance of local context alone, and it even underperforms the global context method on \gpd.
This suggests that, without any additional structure, the self-attention mechanism of the transformer cannot effectively distinguish and leverage information from the local and global contexts.

\begin{table}[t]
    \centering
    \small
    \renewcommand{\arraystretch}{1.0}
    \begin{tabular}{@{}lccc@{}}
    \toprule
    \textbf{Technique}          & \textbf{P@1}       & \textbf{MAP}       & \textbf{MRR}       \\
    \midrule
    N-gram Overlap        & 0.638 & 0.725 & 0.746 \\
    Cosine Similarity     & 0.661 & 0.742 & 0.758 \\
    \bottomrule
\end{tabular}
\caption{Comparison of global context extraction techniques on the ASNQ test set when used with MWA.}
\label{tab:selection}
\end{table}

We note that MWA achieves near identical performance to the ensemble model on both datasets, suggesting that a controlled attention mechanism can overcome limitations in the representation for vanilla transformers, while reducing latency by 21.5\% and memory usage by 89\%.
MWA also matches the latency of the context concatenation model, while improving it by 4.8\% and 3.9\% in P@1 on ASNQ and \gpd, respectively.

Finally, we study the effect of our proposed global extraction techniques in Table~\ref{tab:selection}.
We observe that, among the two proposed algorithms, the cosine similarity approach significantly outperforms the N-gram based method.
This confirms that pre-trained language models can better select context semantically related to question and candidates.

We note n-gram overlap is less computationally taxing, as it can be efficiently implemented as a set of sparse operations over bag of word representations of the question and answer candidates. 
On the other hand, for cosine similarity, it is necessary to compute $\text{Score}_{sim}(C_{i,p} | Q, C_{i,j})$ for all context sentences using a transformer model.
Recently introduced transformer architecture variants could be used to either speed up this similarity computation \citep{cao2020deformer}, or compute query and text representation independently \citep{khattab2020colbert}. 
We leave the evaluation of these techniques to future work.

%% file: sections/conclusions.tex
\section{Conclusions}
\label{sec:conclusions}

For efficiency reasons, traditional \astwo models are designed to estimate answer relevancy by only comparing question and candidates.
In this work, we described and evaluate several techniques to incorporate local and global contexts in the answer selection process.
The results of our experiments show that our proposed methods significantly outperform non-contextual approaches;
further, we empirically demonstrate that local and global contexts can be effectively combined to further improve ranking performance.